\documentclass[11pt,a4paper]{article}
\usepackage[hyperref]{naaclhlt2018}
\usepackage{times}
\usepackage{latexsym}
\usepackage{url}
\usepackage{multirow}
\usepackage{graphicx}
\usepackage[colorinlistoftodos]{todonotes}
\usepackage{booktabs}

\aclfinalcopy 
\title{DpgMedia2019: A Dutch News Dataset for Partisanship Detection}

\author{Chia-Lun Yeh,$^{1,2}$ Babak Loni\thanks{This research was carried out while this author was working at DPG Media.},$^{3}$ Mari\"elle Hendriks,$^{1}$ Henrike Reinhardt,$^{1}$ and Anne Schuth$^{1}$\\
  $^{1}$DPG Media, The Netherlands\\
  $^{2}$TU Delft, The Netherlands\\
  $^{3}$TezolMarket, Iran\\
  {\tt chialunyehlun@gmail.com, babak.loni@tezolmarket.com }\\
  {\tt \{marielle.hendriks, henrike.reinhardt, anne.schuth\}@persgroep.net} }

\date{}

\begin{document}
\maketitle
\begin{abstract}
We present a new Dutch news dataset with labeled partisanship. The dataset contains more than 100K articles that are labeled on the publisher level and 776 articles that were crowdsourced using an internal survey platform and labeled on the article level. In this paper, we document our original motivation, the collection and annotation process, limitations, and applications.
\end{abstract}

\section{Introduction}
In a survey across 38 countries, the Pew Research Center reported that the global public opposed partisanship in news media \cite{pew:2018}. It is, however, challenging to assess the partisanship of news articles on a large scale. We thus made an effort to create a dataset of articles annotated with political partisanship so that content analysis systems can benefit from it.
\newline

To construct a dataset of news articles labeled with partisanship, it is required that some annotators read each article and decide whether it is partisan. This is an expensive annotation process. Another way to derive a label for an article is by using the partisanship of the publisher of the article. Previous work has used this method \cite{potthast:2018, Kulkarni:2018:multiview, kiesel:2019}. This labeling paradigm is premised on that partisan publishers publish more partisan articles and non-partisan publishers publish more non-partisan articles. Although there would be non-partisan articles published by partisan publishers (and vice versa), and thus labeled wrongly, the assumption ensures more information than noise. Once the partisanship of a publisher is known, the labels of all its articles are known, which is fast and cheap. We created a dataset of two parts. The first part contains a large number of articles that were labeled using the partisanship of publishers. The second part contains a few hundreds of articles that were annotated by readers who were asked to read each article and answer survey questions. In the following sections, we describe the collection and annotation of both parts of the dataset.

\section{Dataset description}
\label{sec:dpg}
DpgMedia2019\footnote{The dataset is released at \url{https://github.com/dpgmedia/partisan-news2019}} is a Dutch dataset that was collected from the publications within DPG Media.\footnote{\url{https://www.persgroep.nl/}} We took 11 publishers in the Netherlands for the dataset. These publishers include 4 national publishers, Algemeen Dagblad (AD), de Volkskrant (VK), Trouw, and Het Parool, and 7 regional publishers, de Gelderlander, Tubantia, Brabants Dagblad, Eindhovens Dagblad, BN/De Stem PZC, and de Stentor. The regional publishers are collectively called Algemeen Dagblad Regionaal (ADR). A summary of the dataset is shown in Table \ref{tab:dataset-summary}.

\begin{table*}[t]
\centering
\begin{tabular}{lrr}
\toprule
Label & Publisher-level  & Article-level \\
\midrule
Number of articles & 100K   & 766  \\
Percentage of partisan articles  & 50\%   & 26\%  \\
Number of publishers & 11   & 11   \\
Annotation method & audience-based & crowdsource \\ 
Inter-rater agreement & X  & Krippendorf's alpha: 0.18      \\ \bottomrule
\end{tabular}
\caption{Summary of the two parts of dpgMedia2019 dateset.}
\label{tab:dataset-summary}
\end{table*}

\subsection{Publisher-level data}
We used an internal database that stores all articles written by journalists and ready to be published to collect the articles. From the database, we queried all articles that were published between 2017 and 2019. We filtered articles to be non-advertisement. We also filtered on the main sections so that the articles were not published under the sports and entertainment sections, which we assumed to be less political. After collecting, we found that a lot of the articles were published by several publishers, especially a large overlap existed between AD and ADR. To deal with the problem without losing many articles, we decided that articles that appeared in both AD and its regional publications belonged to AD. Therefore, articles were processed in the following steps:
\begin{enumerate}
    \item Remove any article that was published by more than one national publisher (VK, AD, Trouw, and Het Parool). This gave us a list of unique articles from the largest 4 publishers.
    \item Remove any article from ADR that overlapped with the articles from national publishers.
    \item Remove any article that was published by more than one regional publisher (ADR).
\end{enumerate}
The process assured that most of the articles are unique to one publisher. The only exceptions were the AD articles, of which some were also published by ADR. This is not ideal but acceptable as we show in the section \ref{sec:publisher_partisanship} that AD and ADR publishers would have the same partisanship labels. In the end, we have 103,812 articles.

\subsubsection{Annotation of publisher partisanship}
\label{sec:publisher_partisanship}
To our knowledge, there is no comprehensive research about the partisanship of Dutch publishers. We thus adopted the audience-based method to decide the partisanship of publishers. Within the survey that will be explained in section \ref{subsec:dpgMedia2019_annotation}, we asked the annotators to rate their political leanings. The question asked an annotator to report his or her political standpoints to be extreme-left, left, neutral, right, or extreme-right. We mapped extreme-left to -2, left to -1, center to 0, right to 1, extremely-right to 2, and assigned the value to each annotator. Since each annotator is subscribed to one of the publishers in our survey, we calculated the partisanship score of a publisher by averaging the scores of all annotators that subscribed to the publisher. The final score of the 11 publishers are listed in Table \ref{tab:dpgMedia2019-publisher-partisanship}, sorted from the most left-leaning to the most right-leaning.
\begin{table}[t]
    \centering
    \begin{tabular}{lrr}
    \toprule
     Publisher & \#annotators & Score  \\
     \midrule
     vk & 780 & -0.6372\\
     trouw & 491 & -0.5438\\
     parool & 410 & -0.4341\\
     degelderlander & 208 & -0.2452\\
     tubantia & 236 & -0.1864\\
     drabantsdagblad & 188 & -0.1862\\
     eindhovensdagblad & 194 & -0.0722\\
     destem & 151 & -0.0464\\
     pzc & 202 & -0.0248\\
     ad & 695 & 0.0302\\
     destentor & 55 & 0.0727\\
     \midrule
     & total: 3,610 & mean: -0.2067\\
     \bottomrule
\end{tabular}
    \caption{Publisher, number of people of whom the political leaning we know, and the computed partisanship score.}
    \label{tab:dpgMedia2019-publisher-partisanship}
\end{table}

We decided to treat VK, Trouw, and Het Parool as partisan publishers and the rest non-partisan. This result largely accords with that from the news media report from the Pew Research Center in 2018 \cite{pew-nl:2018}, which found that VK is left-leaning and partisan while AD is less partisan. 

Table \ref{tab:filter_article_length} shows the final publisher-level dataset of dpgMedia2019, with the number of articles and class distribution.
\begin{table*}[h!]
    \centering
\begin{tabular}{@{}lcccccc@{}}
\toprule
Partisanship & \multicolumn{3}{c}{Partisan} & &\multicolumn{2}{c}{Non-partisan} \\ 
\cmidrule{2-4} \cmidrule{6-7}
Publisher & de Volkskrant & Trouw & Het Parool && AD & ADR \\
Article Num. &  11,761 & 21,614 &  19,498 &&  40,029   &  10,910 \\
\midrule
Total & \multicolumn{3}{c}{52,873 (50.9\%)} & &\multicolumn{2}{c}{50,939 (49.1\%)} \\
\bottomrule
\end{tabular}
    \caption{Number of articles per publisher and class distribution of publisher-level part of dpgMedia2019.}
    \label{tab:filter_article_length}
\end{table*}

\subsection{Article-level data}
\label{subsec:dpgMedia2019_annotation}
To collect article-level labels, we utilized a platform in the company that has been used by the market research team to collect surveys from the subscribers of different news publishers. The survey works as follows: The user is first presented with a set of selected pages (usually 4 pages and around 20 articles) from the print paper the day before. The user can select an article each time that he or she has read, and answer some questions about it. We added 3 questions to the existing survey that asked the level of partisanship, the polarity of partisanship, and which pro- or anti- entities the article presents. We also asked the political standpoint of the user. The complete survey can be found in Appendices.
\newline

The reason for using this platform was two-fold. First, the platform provided us with annotators with a higher probability to be competent with the task. Since the survey was distributed to subscribers that pay for reading news, it's more likely that they regularly read newspapers and are more familiar with the political issues and parties in the Netherlands. On the other hand, if we use crowdsourcing platforms, we need to design process to select suitable annotators, for example by nationality or anchor questions to test the annotator's ability. Second, the platform gave us more confidence that an annotator had read the article before answering questions. Since the annotators could choose which articles to annotate, it is more likely that they would rate an article that they had read and had some opinions about.
\newline

The annotation task ran for around two months in February to April 2019. We collected annotations for 1,536 articles from 3,926 annotators. 
\subsubsection{Annotation distributions}
For the first question, where we asked about the intensity of partisanship, more than half of the annotations were non-partisan. About 1\% of the annotation indicated an extreme partisanship, as shown in Table \ref{tab:dpgMedia2019-partisanship-intensity}. For the polarity of partisanship, most of the annotators found it not applicable or difficult to decide, as shown in Table \ref{tab:dpgMedia2019-partisanship-polartiy}. For annotations that indicated a polarity, the highest percentage was given to progressive. Progressive and conservative seemed to be more relevant terms in the Netherlands as they are used more than their counterparts, left and right, respectively.
\begin{table*}[h!]
    \centering
    \begin{tabular}{@{}cccccc@{}}
    \toprule
    non-partisan & \begin{tabular}[c]{@{}l@{}}reasonably non-\\partisan\end{tabular} & \begin{tabular}[c]{@{}l@{}}somewhat\\partisan\end{tabular} & partisan & \begin{tabular}[c]{@{}l@{}}extremely\\partisan\end{tabular} & \begin{tabular}[c]{@{}l@{}}impossible\\to decide\end{tabular}\\ \midrule
    52.85\% & 16.34\% & 10.54\% & 5.49\% & 0.91\% & 13.88\% \\
    \bottomrule
    \end{tabular}
    \caption{Distribution of annotations of the strength of partisanship.}
    \label{tab:dpgMedia2019-partisanship-intensity}
\end{table*}

\begin{table*}[h!]
    \centering
    \begin{tabular}{@{}ccccccc@{}}
    \toprule
        left & right & progressive & conservative & others & not applicable & unknown\\ \midrule
        5.66\% & 2.74\% & 7.74\% & 2.78\% & 7.29\% & 54.81\% & 18.96\%\\
        \bottomrule
    \end{tabular}
    \caption{Distribution of annotations of the polarity of partisanship.}
    \label{tab:dpgMedia2019-partisanship-polartiy}
\end{table*}

As for the self-rated political standpoint of the annotators, nearly half of the annotators identified themselves as left-leaning, while only around 20\% were right-leaning. This is interesting because when deciding the polarity of articles, left and progressive ratings were given much more often than right and conservative ones. This shows that these left-leaning annotators were able to identify their partisanship and rate the articles accordingly. 
\begin{table*}[h!]
    \centering
    \begin{tabular}{@{}ccccc@{}}
    \toprule
        extreme-left & left & middle & right & extreme-right\\ \midrule
        1.14\% & 46.87\% & 32.71\% & 19.14\% & 0.14\% \\
        \bottomrule
    \end{tabular}
    \caption{Distribution of annotations of self-identified political standpoints.}
    \label{tab:annotator-political-leaning}
\end{table*}

\begin{figure*}[h!]
    \centering
    \includegraphics[width = 0.8\textwidth]{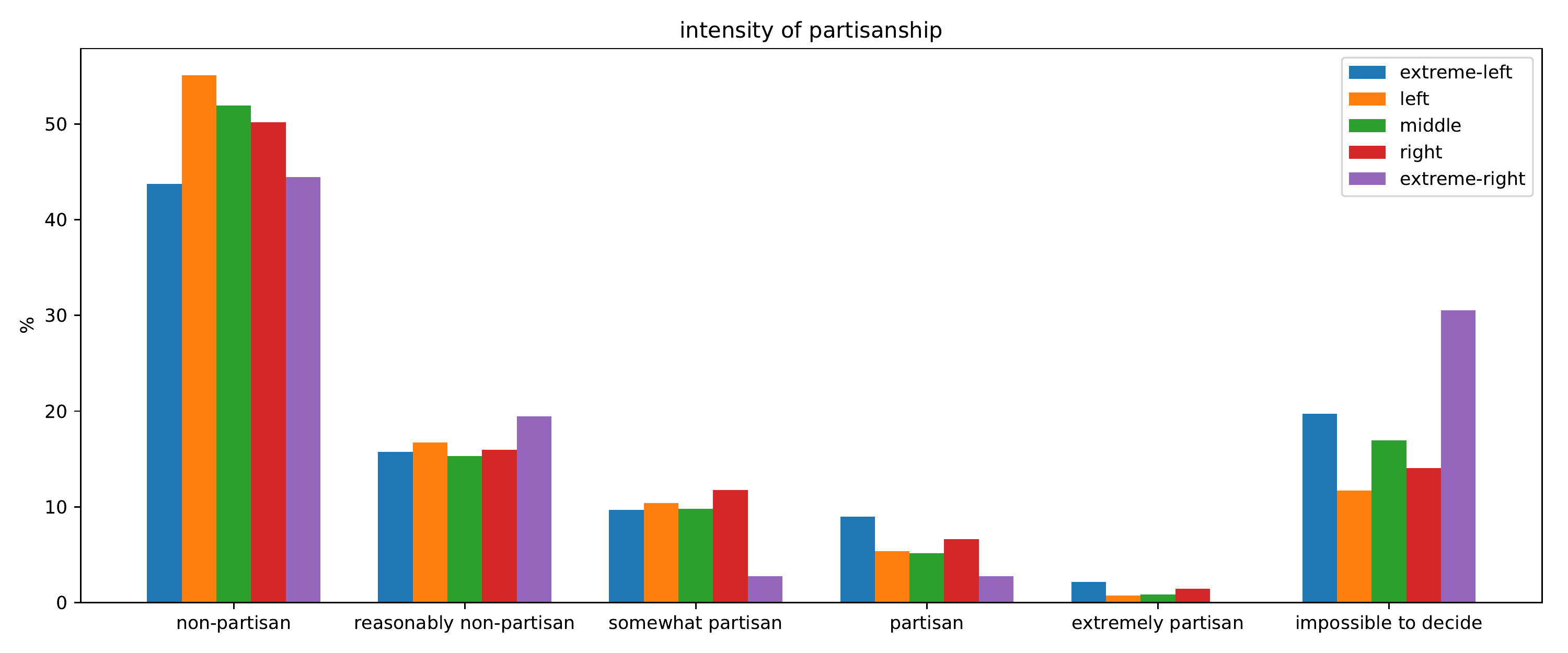}
    \caption{Percentage of annotation grouped by political leaning and annotation for the intensity of partisanship.}
    \label{fig:annotation-per-group-q1}
\end{figure*}

\begin{figure*}[h!]
    \centering
    \includegraphics[width=0.8\textwidth]{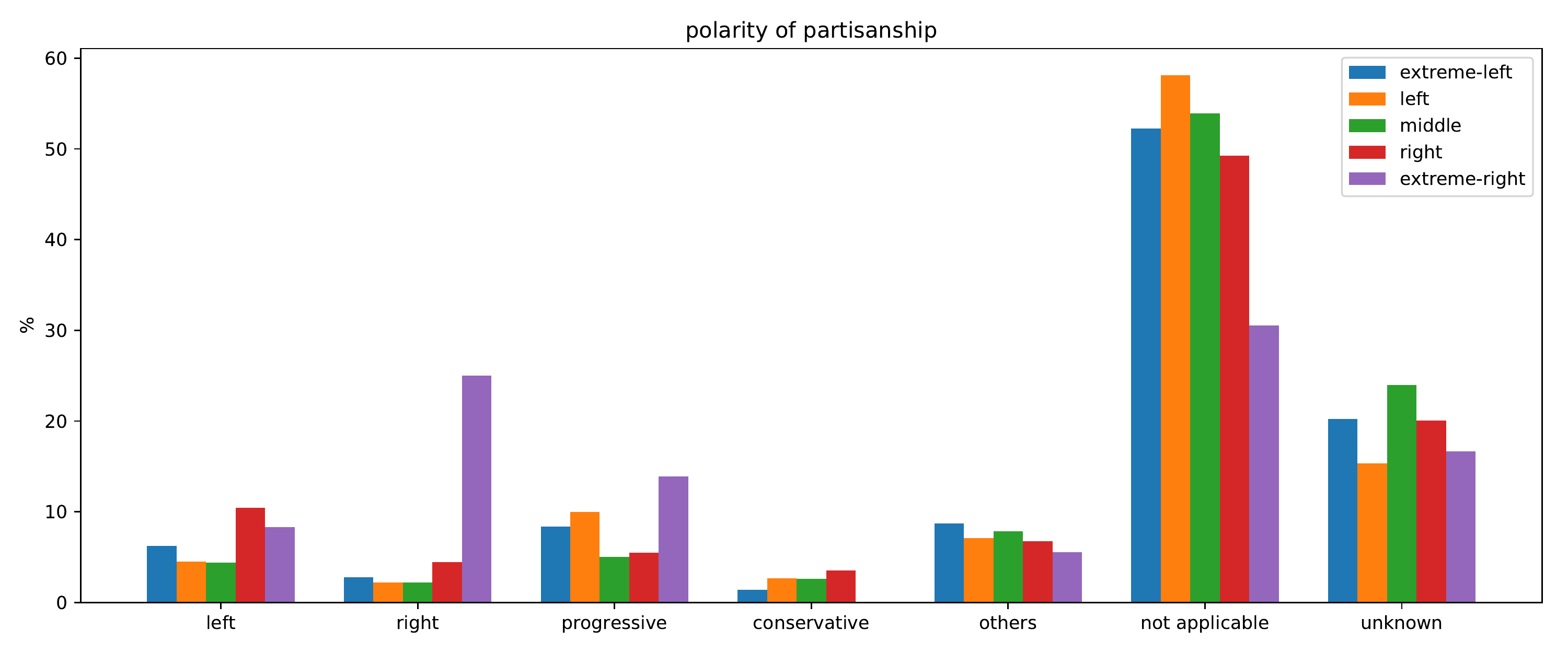}
    \caption{Percentage of annotation grouped by political leaning and annotation for the polarity of partisanship.}
    \label{fig:annotation-per-group-q2}
\end{figure*}

We suspected that the annotators would induce bias in ratings based on their political leaning and we might want to normalize it. To check whether this was the case, we grouped annotators based on their political leaning and calculate the percentage of each option being annotated. In Figure \ref{fig:annotation-per-group-q1}, we grouped options and color-coded political leanings to compare whether there are differences in the annotation between the groups. We observe that the "extreme-right" group used less "somewhat partisan", "partisan", and "extremely-partisan" annotations. This might mean that articles that were considered partisan by other groups were considered "non-partisan" or "impossible to decide" by this group. We didn't observe a significant difference between the groups. 
Figure \ref{fig:annotation-per-group-q2} shows the same for the second question. Interestingly, the "extreme-right" group gave a lot more "right" and slightly more "progressive" ratings than other groups. In the end, we decided to use the raw ratings. How to scale the ratings based on self-identified political leaning needs more investigation.


\subsubsection{Quality control and agreement analysis}
The main question that we are interested in is the first question in our survey. In addition to the 5-point Likert scale that an annotator could choose from (non-partisan to extremely partisan), we also provided the option to choose "impossible to decide" because the articles could be about non-political topics. When computing inter-rater agreement, this option was ignored. The remaining 5 ratings were treated as ordinal ratings. The initial Krippendorff's alpha was 0.142, using the interval metric. To perform quality control, we devised some filtering steps based on the information we had. These steps are as follows:
\begin{enumerate}
    \item Remove uninterested annotators: we assumed that annotators that provided no information were not interested in participating in the task. These annotators always rated "not possible to decide" for Q1, 'not applicable' or "unknown" for Q2, and provide no textual comment for Q3. There were in total 117 uninterested annotators and their answers were discarded.
    \item Remove unreliable annotators: as we didn't have "gold data" to evaluate reliability, we used the free text that an annotator provided in Q3 to compute a reliability score. The assumption was that if an annotator was able to provide texts with meaningful partisanship description, he or she was more reliable in performing the task. To do this, we collected the text given by each annotator. We filtered out text that didn't answer the question, such as symbols, 'no idea', 'see above', etc. Then we calculated the reliability score of annotator $i$ with equation \ref{equa:score}, where $t_{i}$ is the number of clean texts that annotator $i$ provided in total and $N_{i}$ is the number of articles that annotator $i$ rated.
    \begin{equation}
    \label{equa:score}
        score_{i} = \frac{t_{i} + 1}{N_{i}} \times (t_{i}+1)
    \end{equation}
    We added one to $t_{i}$ so that annotators that gave no clean texts would not all end up with a zero score but would have different scores based on how many articles they rated. In other words, if an annotator only rated one article and didn't give textual information, we considered he or she reliable since we had little information. However, an annotator that rated ten articles but never gave useful textual information was more likely to be unreliable.
    The reliability score was used to filter out annotators that rarely gave meaningful text. The threshold of the filtering was decided by the Krippendorff's alpha that would be achieved after discarding the annotators with a score below the threshold.
    \item Remove articles with too few annotations: articles with less than 3 annotations were discarded because we were not confident with a label that was derived from less than 3 annotations. 
    \item Remove unreliable articles: if at least half of the annotations of an article were "impossible to decide", we assumed that the article was not about issues of which partisanship could be decided. 
\end{enumerate}

Finally, we mapped ratings of 1 and 2 to non-partisan, and 3 to 5 to partisan. A majority vote was used to derive the final label. Articles with no majority were discarded. In the end, 766 articles remained, of which 201 were partisan. Table \ref{tab:dpgMedia2019-article-partisanship} shows the number of articles and the percentage of partisan articles per publisher. The final alpha value is 0.180.
\begin{table}[h!]
    \centering
    \begin{tabular}{lrr}
    \toprule
     Publisher & \#articles & \%partisan\\
     \midrule
     vk & 166 & 27.11\\
     trouw & 140 & 25.00\\
     parool & 121 & 28.93\\
     degelderlander & 46 &19.57\\
     tubantia & 34 & 41.18\\
     brabantsdagblad & 32 & 31.25\\
     eindhovensdagblad & 20 & 35.00\\
     destem& 34 & 17.65\\
     pzc & 30& 26.67\\
     ad & 133 & 24.06\\
     destentor & 10 & 0.00\\
     \midrule
     & total: 766 & mean: 26.24\\
     \bottomrule
\end{tabular}
    \caption{Number of articles and percentage of partisan articles by publisher.}
    \label{tab:dpgMedia2019-article-partisanship}
\end{table}

\section{Analysis of the datasets}
\label{sec:analysis}
In this section, we analyze the properties and relationship of the two parts (publisher-level and article-level) of the datasets. In Table \ref{tab:statistics}, we listed the length of articles of the two parts. The reason that this is important is to check whether there are apparent differences between the articles in the two parts of the dataset. We see that the lengths are comparable, which is desired.

\begin{table}[h!]
    \centering
    \begin{tabular}{lrr}
    \toprule
    article length & publisher-level & article-level\\ \midrule
    Mean &  470.1 & 471.2\\
    SD &  387.5 & 275.1\\
    50\% percentile & 381.0 & 451.0\\ \bottomrule
    \end{tabular}
    \caption{Statistics of length of articles.}
    \label{tab:statistics}
\end{table}

The second analysis is the relationship between publisher and article partisanship. We want to check whether the assumption of partisan publishers publish more partisan articles is valid for our dataset. To do this, we used the article-level labels and calculated the percentage of partisan articles for each publisher. This value was then compared with the publisher partisanship. We calculated Spearsman's correlation between the publisher partisanship derived from the audience and article content. We take the absolute value of the partisanship in table \ref{tab:dpgMedia2019-publisher-partisanship} and that in table  \ref{tab:dpgMedia2019-article-partisanship}. The correlation is 0.21. This low correlation resulted from the nature of the task and publishers that were considered. The partisan publishers in DPG Media publish news articles that are reviewed by professional editors. The publishers are often partisan only on a portion of the articles and on certain topics.

\section{Limitations}
We identified some limitations during the process, which we describe in this section.

When deciding publisher partisanship, the number of people from whom we computed the score was small. For example, de Stentor is estimated to reach 275K readers each day on its official website. Deciding the audience leaning from 55 samples was subject to sampling bias. Besides, the scores differ very little between publishers. None of the publishers had an absolute score higher than 1, meaning that even the most partisan publisher was only slightly partisan. Deciding which publishers we consider as partisan and which not is thus not very reliable.

The article-level annotation task was not as well-defined as on a crowdsourcing platform. We included the questions as part of an existing survey and didn't want to create much burden to the annotators. Therefore, we did not provide long descriptive text that explained how a person should annotate an article. We thus run under the risk of annotator bias. This is one of the reasons for a low inter-rater agreement.

\section{Dataset Application}
This dataset is aimed to contribute to developing a partisan news detector. There are several ways that the dataset can be used to devise the system. For example, it is possible to train the detector using publisher-level labels and test with article-level labels. It is also possible to use semi-supervised learning and treat the publisher-level part as unsupervised, or use only the article-level part. We also released the raw survey data so that new mechanisms to decide the article-level labels can be devised.

\section{Acknowledgements}
We would like to thank Johannes Kiesel and the colleagues from Factmata for providing us with the annotation questions they used when creating a hyperpartisan news dataset. We would also like to thank Jaron Harambam, Judith M\"oller for helping us in asking the right questions for our annotations and Nava Tintarev for sharing her insights in the domain. 

\appendixpage
\label{appendix}
We list the questions we asked in the partisanship annotation survey, in the original Dutch language and an English translation. 
\begin{itemize}
    \item \textbf{Q1}: Over het algemeen, hoe bevooroordeeld vindt u dit artikel? Een artikel dat bevooroordeeld is, is voor of tegen een persoon of groep. N.B. Een artikel kan gaan over controversiële onderwerpen, zoals politiek, maar blijft redelijk neutraal.
    \begin{enumerate}
        \item Onbevooroordeeld
        \item Redelijk onbevooroordeeld
        \item Enigszins bevooroordeeld
        \item Bevooroordeeld
        \item Extreem bevooroordeeld
        \item Onmogelijk om te bepalen
    \end{enumerate}
    \item \textbf{Q2}: Als u vindt dat dit artikel bevooroordeeld is, ten gunste van welke politieke richting vindt u dit artikel geschreven? (U kunt meerdere antwoorden kiezen)
    \begin{enumerate}
        \item Links
        \item Rechts
        \item Progressief
        \item Conservatief
        \item Anders, namelijk: OPEN
        \item Niet van toepassing op dit artikel
        \item Ik weet het niet
    \end{enumerate}
    \item \textbf{Q3}: Als u vindt dat het artikel bevooroordeeld is, pro of anti wie of wat vindt u dit artikel? (bijvoorbeeld pro-PVV, pro-conservatieven, pro-kapitalist, anti-Trump, anti-moslims, anti-atheïst). (U kunt meerdere antwoorden kiezen)
    \begin{itemize}
        \item Pro:
        \item Anti:
    \end{itemize}
    \item \textbf{Q4}: Hoe zou u uw eigen politieke standpunt bepalen?
    \begin{enumerate}
        \item Extreemlinks
        \item (gematigd-)links
        \item Neutraal
        \item (gematigd-)rechts
        \item Extreemrechts
    \end{enumerate}
\end{itemize}

\textbf{Translated}
\begin{itemize}
    \item \textbf{Q1}: Overall, how biased is this article? An article that is biased is for or against a person or group. Note that an article can talk about contentious topics, like politics, but remains fairly neutral.
    \begin{enumerate}
        \item Unbiased
        \item Fairly unbiased
        \item Somewhat biased
        \item Biased
        \item Extremely biased
        \item Not possible to decide
    \end{enumerate}
    
    \item \textbf{Q2}: If you find the article biased, which political direction do you find this article in favor of? (You can choose multiple answers)
    \begin{enumerate}
        \item Left
        \item Right
        \item Progressive
        \item Conservative
        \item Others
        \item Not applicable to the article
        \item I don't know
    \end{enumerate}
    
    \item \textbf{Q3}: If you find the article biased, indicate who or what the article is biased in favor of ('pro') and/or against ('anti')? (for example pro-PVV, pro-conservative, pro-capitalist, anti-Trump, anti-Muslims, anti-atheist). (You can have multiple answers)
    \begin{itemize}
        \item Pro:
        \item Anti:
    \end{itemize}
    \item \textbf{Q4}: How would you determine your own political position?
    \begin{enumerate}
        \item Extreme-left
        \item (moderate)left
        \item Neutral
        \item (moderate)right
        \item Extreme-right
    \end{enumerate}
\end{itemize}

\bibliography{semeval2018}
\bibliographystyle{acl_natbib}

\end{document}